\documentclass{article}
\usepackage{kr}

\usepackage{times}
\usepackage{helvet}  
\usepackage{courier}  
\usepackage{soul}
\usepackage{url}
\usepackage[hidelinks]{hyperref}
\usepackage[utf8]{inputenc}
\usepackage[small]{caption}
\usepackage{graphicx}
\usepackage{amsmath}
\usepackage{amsthm}
\usepackage{booktabs}
\usepackage{algorithm,algpseudocode}
\usepackage{algorithmicx}
\usepackage{mathtools}

\usepackage{latexsym}
\usepackage{amsfonts,amsmath,amssymb,amsthm}

\usepackage{newfloat}
\usepackage{listings} 
\DeclareCaptionStyle{ruled}{labelfont=normalfont,labelsep=colon,strut=off} 
\floatstyle{ruled}
\newfloat{listing}{tb}{lst}{}
\floatname{listing}{Listing}
\newcommand{\imply}{\rightarrow}
\newcommand{\liff}{\leftrightarrow}

\newcommand{\eq}{\!=\!}

\newcommand{\lleq}{\!\leq\!}
\newcommand{\lland}{\!\land\!}
\newcommand{\llor}{\!\lor\!}

\newcommand{\commentout}[1]{}

\newcommand{\isdef}{\hbox{$\stackrel{{\scriptstyle def}}{=}$}}
\newcommand{\sa}{S_{\alpha}}
\newcommand{\bat}{\mathcal{D}}
\newcommand{\dsz}{\mathcal{D}_{S_0}}
\newcommand{\dsa}{\mathcal{D}_{\sa}}

\newcommand{\movebtob}{\mbox{\it move-b-to-b}}
\newcommand{\movebtot}{\mbox{\it move-b-to-t}}
\newcommand{\movettob}{\mbox{\it move-t-to-b}}

\mathchardef\mhyphen="2D

\renewcommand{\algorithmicrequire}{\textbf{Input: }}
\renewcommand{\algorithmicensure}{\textbf{Output: }}

\algnewcommand\algorithmicdefn{\textbf{Definitions:}}
\algnewcommand\Defn{\item[\algorithmicdefn]}

\newtheorem{theorem}{Theorem}

\theoremstyle{definition}

\renewcommand{\titlebox}{1.4in}

\title{Planning as Theorem Proving with Heuristics}
\author{
Mikhail Soutchanski$^1$\and
Ryan Young$^2$\\
\affiliations
$^{(1,2)}$Toronto Metropolitan University, 245 Church St, ENG281, Toronto, ON, M5B 2K3, Canada
}

\begin{document}

\maketitle

\begin{abstract}
Planning as theorem proving in situation calculus was abandoned 50 years ago as an impossible project. 
But we have developed a Theorem Proving Lifted Heuristic (TPLH) planner that 
searches for a plan in a tree of situations using the A* search algorithm.
It is controlled by a delete relaxation-based domain independent heuristic.
We compare TPLH with Fast Downward (FD) and Best First Width Search (BFWS) planners 
over several standard benchmarks. Since our implementation of the heuristic function 
is not optimized, TPLH is slower than FD and BFWS. But it computes shorter plans, 
and it explores fewer states. 
We discuss previous research on planning within KR\&R and identify related directions. 
Thus, we show that deductive lifted heuristic 
planning in situation calculus is actually doable.
\end{abstract}

\section{Introduction}
From the very beginning of Artificial Intelligence (AI), it was recognized that 
the planning problems at the common sense level can be conveniently formulated 
in situation calculus (SC) as the entailment problem in predicate logic.
\cite{McCarthy1963}.  
In 1969, Green proposed to solve the (planning) problems 
using resolution-based theorem proving in situation calculus
\cite{Green1969,GreenPhD1969}.
However, because at that time the resolution algorithms were very inefficient,
the famous frame problem was not yet solved even for direct effects, 
and because he did not anticipate any control over unrestricted resolution, 
his computer program did not work well, and there was no immediate fix. 
This is discussed in many publications, e.g., see
\cite{Hayes1973,Raphael1976,KowalskiCommACM1988,WeldAIMag1999,Kowalski2014}. 

Subsequent planning research moved away from theorem proving in SC to specialized 
non-logical representations such as STRIPS \cite{FikesNilsson1971} and ADL \cite{Pednault1994}.
In modern automated (common-sense) AI planning, the instances of 
the planning problem are usually solved with domain-independent heuristics 
\cite{GeffnerBonetBook2013,GhallabNauTraverso2004} in a single model of 
a discrete transition system using model-based approaches. 
Meanwhile, the frame problem for direct effects of actions has been solved for
a large class of action theories \cite{Reiter1991,Reiter2001}, and the SLD resolution
algorithms were efficiently implemented, helping to develop PROLOG 
\cite{Colmerauer1990,ColmerauerRoussel1996,VanEmdenKowalski1976,CohenTribute2001,Lloyd2012,Prolog50years}. 

There were several proposals to continue Green's planning as theorem proving approach, e.g., 
\cite{BibelNGC1986,BibelAI1998,MannaWaldinger1987,HolldoblerSchneebergerNGC1990,BiundoEWSP1993,Fronhofer1996,Fronhfer1997Inform,LevesqueAAAI1996,LinAMAI1997,MuellerSutcliffeLPAR2005,SchiffelThielscher2006,LiLiuAAAI2020}, 
and implement the automated deterministic planners in PROLOG e.g., see  
\cite{ShanahanJLP2000,FinziPirriReiterOWA2000,LinPlannerR,ZhouBartakDovier2015,BatusovMScThesis2014,QovaiziMScThesis2019}, 
or compute plans using the Answer Set Programming, e.g., see 
\cite{LifschitzICLP1999,SonPontelliBalducciniSchaub2022}, 
but they were not supported with efficient 
implementations using domain independent heuristics. 
Despite this history, there are potential advantages to theorem-proving based 
planning over the model-based approach, e.g., the former can operate even if 
there are many different infinite logical models \cite{Enderton} for
an application domain, or if an initial state theory is incomplete. 
Model-based approaches rely on both the Closed World Assumption (CWA) and 
the Domain Closure Assumption (DCA)
\cite{Reiter1977,Reiter1978,DBLP:journals/jacm/Reiter80}. 
The main reason for relying on the (unrealistic) DCA in model-based approaches 
is the need for instantiating the transition system before search starts. 
But the DCA can be avoided in a theorem-proving based forward search planner 
using progression in local effect SC action theories \cite{LiuLakemeyer2009} 
with an incomplete initial theory (i.e. no CWA) \cite{LakemeyerLevesqueKR2002}. 

In this paper, we revisit the deductive approach to planning in SC despite a 
common incorrect belief that efficient deductive planning in SC is impossible.
In particular, we show how to design a new SC-based Theorem Proving Lifted 
Heuristic (TPLH) planner that builds on \cite{Soutchanski2017}. It is ``lifted" 
in a sense that it works with action schemata at run-time, but not with actions 
instantiated before planning starts. Our planner does forward search over a situation 
tree, in contrast to modern model-based planners that usually search over a state space.
Moreover, it makes use of a domain independent heuristic. To the best of our knowledge 
this is the first ever deductive planner to incorporate both of these features.
This is our main contribution.
Essentially, our $A^*$ planner provides heuristic control over resolution, but 
in \cite{Green1969,GreenPhD1969} control was not anticipated. 
The current version of our planner works with a domain 
independent delete relaxation heuristic inspired by the FF planner
\cite{HoffmannNebel2001,BryceKambhampati2007}, 
but any other domain independent heuristics can be implemented as well.

We start with a review of SC, then 
we explain how our TPLH planner can be developed in SC from the first principles
as a search over the situation tree. 
We did experimental comparison with the state-of-the-art (SOTA) planners. 
We compare the performance of our TPLH planner, the recent version of FastDownward (FD) planner
\cite{HelmertJAIR2006,DBLP:journals/jair/RichterW10,FD} and 
Best First Width Search (BFWS) planner \cite{DBLP:conf/aaai/LipovetzkyG17,BFWS}
on a set of the usual PDDL \cite{DBLP:series/synthesis/2019Haslum}
benchmarks. We report the experimental results, 
discuss all known directly related previous work, future research directions 
and then conclude.

\section{Background}
The situation calculus (SC) is a logical approach to representation and 
reasoning about actions and their effects. It was introduced in 
\cite{McCarthy1963,McCarthyHayes1969}
to capture common sense reasoning about the actions and events that can change 
properties of the world and mental states of the agents. SC was refined by 
Reiter \cite{Reiter1991,Reiter2001} who introduced the Basic Action Theories (BAT).
Unlike the notion of state that is common in model-based planning, 
SC relies on situation, namely a sequence of actions, 
which is a concise symbolic representation and a convenient proxy 
for the state in the cases when all actions are deterministic 
\cite{LevesquePirriReiter1998,LinKRhandbook2008}. 
We use variables $s,s',s_1,s_2$ for situations, variables $a,a'$ for actions,
and $\bar{x},\bar{y}$ for tuples of object variables. 
The constant $S_0$ represents the initial situation, and the successor function 
$do: action\times situation \mapsto situation$, e.g., $do(a,s)$, denotes 
situation that results from doing action $a$ in previous situation $s$. 
The terms $\sigma,\sigma'$ denote situation terms, and
$A_i(\bar{x})$, or $\alpha,\alpha_1,\alpha_2,\alpha'$, represent 
action functions and action terms, respectively. 
The shorthand $do([\alpha_1,\cdots,\alpha_n],S_0))$
represents situation $do(\alpha_n,do(\cdots, do(\alpha_1,S_0)\cdots))$
resulting from execution of actions $\alpha_1,\cdots,\alpha_n$ in $S_0$.
The relation $\sigma \sqsubset \sigma'$ between situations terms $\sigma$ and 
$\sigma'$ means that $\sigma$ is an initial sub-sequence of $\sigma'$. 
Any predicate symbol $F(\bar{x},s)$
with exactly one situation argument $s$ and possibly a tuple of object arguments
$\bar{x}$ is called a (relational) fluent. Without loss of generality, 
we consider only relational fluents in this paper, but the language of SC 
can also include functional fluents. A first order logic (FO) formula $\psi(s)$
composed from fluents, equalities and situation independent predicates is called
\textit{uniform in $s$} if all fluents in the formula $\psi$ mention only situation $s$ 
as their situation argument, and there are no quantifiers over $s$ in the formula.

The basic action theories (BAT)${\cal D}$ is the conjunction of the following 
classes of axioms 
$
{\cal D}\!=\!\Sigma \land {\cal D}_{ss} \land {\cal D}_{ap} \land {\cal D}_{una}\land {\cal D}_{S_0}$\\
We use examples from the Blocks World (BW) application 
domain \cite{Reiter2001,CookLiu2003}. For brevity, all \ $\bar{x},a,s$ 
variables are assumed $\forall$-quantified at the outer level.

$\mathbf{{\cal D}_{ap}}$ is a set of action precondition axioms of the form\\
$\hspace*{0.5in} 
\forall s\forall \bar{x}. \ poss(A(\bar{x}), s)\liff \Pi_A(\bar{x},s),$\\
where $poss(a,s)$ is a new predicate symbol meaning that an action $a$ is 
possible in situation $s$,  $\Pi_A(\bar{x},s)$ is a formula  uniform in $s$, 
and $A$ is an n-ary action function. 
In most planning benchmarks, the formula $\Pi_A$ is simply 
a conjunction of fluent literals and possibly negations of equality. 
We consider a version of BW, where there are three actions: 
$\movebtob(x,y,z)$, move a block $x$ from a block $y$ to another block $z$, 
$\movebtot(x,y)$, move a block $x$ from a block $y$ to the table, 
$\movettob(x,z)$, move a block $x$ from the table to a block $z$.
\\
$
\hspace{-2mm}
\begin{array}{l}
poss(\movebtob(x,\!y,\!z),s) \liff  clear(x,\!s)\land clear(z,\!s)\land\\
\hspace{1.5in}
         on(x,y,s)\land x \not= z.\\
poss(\movebtot(x,y),s) \liff  clear(x,s)\land on(x,y,s).\\
poss(\movettob(x,z),s) \liff  ontable(x,s)\land \\
\hspace{1.5in}     clear(x,s)\land clear(z,s).
\end{array}
$\\

Let $\mathbf{{\cal D}_{ss}}$ be a set of the successor state axioms (SSA):\\
$
\begin{array}{c}
F(\bar{x}, do(a,\!s)) \leftrightarrow \gamma_F^{+}(\bar{x}, a,\! s) \lor \ 
	F(\bar{x}, s) \land \neg \gamma_F^{-}(\bar{x}, a,\! s),
\end{array}\\[0.5ex]
$
\noindent where $\bar{x}$ is a tuple of object arguments of the fluent $F$, and
each of the $\gamma_F$'s is a disjunction of uniform formulas
\ $[\exists\bar{z}]. a = A(\bar{u}) \land \phi(\bar{x},\bar{z},s)$,
where $A(\bar{u})$ is an action with a tuple $\bar{u}$ of object arguments, 
$\phi(\bar{x},\bar{z},s)$ is a context condition, and 
$\bar{z}\subseteq \bar{u}$ are optional object arguments. 
It may be that $\bar{x}\subset \bar{u}$.
%
If $\bar{u}$ in an action function $A(\bar{u})$ does not include any $z$  
variables, then there is no optional $\exists\bar{z}$  quantifier.
If not all variables from $\bar{x}$ are included in $\bar{u}$, then it is said 
that $A(\bar{u})$ has a {\it global effect}, since the fluent $F$ has 
at least one $\forall$-quantified object argument $x$ not included in $\bar{u}$. 
Therefore, $F$ experiences changes beyond the objects explicitly named in $A(\bar{u})$.
For example, if a truck drives from one location to another, driving action
does not mention any boxes loaded on the truck, then the location of {\it all} loaded
boxes change. When the tuple of action arguments $\bar{u}$ contains all fluent
arguments $\bar{x}$, and possibly contains $\bar{z}$, we say that the action 
$A(\bar{u})$ has a {\it local effect}. 
A BAT is called a {\it local-effect} BAT if all of its actions have only local 
effects. In a local-effect action theory, each action can change values of 
fluents only for objects explicitly named as arguments of the action. In our 
implementation, we focus on a simple class of local-effect BAT, where SSAs have no
context conditions. However, since \cite{Pednault1994}, it is common to consider 
a broader class of SSAs with conditional effects that depend on contexts 
$\phi(\bar{x},\bar{z},s)$. Often, contexts are quantifier-free formulas, 
and then SSA is called {\it essentially quantifier-free}. 
In BW, we consider fluents 
$clear(x, s)$, block $x$ has no blocks on top of it,
$on(x, y, s)$, block $x$ is on block $y$ in situation $s$, 
$ontable(x, s)$, block $x$ is on the table is $s$. 
The following SSAs are local-effect (with implicit outside $\forall x,\forall y,\forall a,\forall s$).
$
\begin{array}{l}
clear(x, do(a, s)) \leftrightarrow \\
     \hspace*{-1mm}   \exists y,\!z(a\! =\! \movebtob(y,\! x,\! z))\llor
                    \exists y(a\!=\! \movebtot(y,\!x)) \lor\\ 
     \hspace*{10mm} 
     clear(x, s) \land \neg \exists y, z(a\! =\! \movebtob(y, z, x)) \land\\
          \hspace*{25mm} 
                 \neg \exists y(a\!=\! \movettob(y,x)),\\
on(x, y, do(a, s)) \leftrightarrow \\
      \hspace*{-1mm}  \exists z(a\! =\! \movebtob(x, z, y)) \llor 
                      \exists y(a\!=\! \movettob(x,y) \lor\\
        \hspace*{10mm}
    on(x, y, s) \land \neg\exists z(a\! =\! \movebtob(x, y, z)) \land\\
          \hspace*{25mm} 
                 \neg \exists y(a\!=\! \movebtot(x,y)),\\
ontable(x, do(a,s)) \liff \exists y(a\!=\! \movebtot(x,y)) \lor\\
        \hspace*{10mm}
    ontable(x,s) \land \neg \exists y(a\!=\! \movettob(x,y)).\\
\end{array}\\
$

  ${\cal D}_{una}$ is a finite set of unique name axioms (UNA) for actions and
named objects.
For example,\\  
$
\begin{array}{c}
     \movebtob(x,y,z)\not= \movebtot(x,y),\\
\movebtot(x,y)\!=\!\movebtob(x',y')\imply
            x\!=\!x'\land y\!=\!y'.\\
\end{array}
$

${\cal D}_{S_0}$ is a set of FO formulas whose only situation term is $S_0$. 
It specifies the values of fluents in the initial state. It describes all 
the (static) facts that are not changeable by actions. 
Also, it includes {\it domain closure for actions} such as 
$
\begin{array}{l}
\forall a. \ \exists x,y,z(a\eq \movebtob(x,y,z))\ \lor \\
            \hspace{10mm}   \exists x,y(a\eq \movebtot(x,y) )\, \lor\\
            \hspace{20mm} \exists x,y(a\eq \movettob(x,y) ).
\end{array}
$\\
In particular, it may include axioms for domain specific constraints 
(state axioms), e.g.,\\
$
\begin{array}{l}
\forall x\forall y( on(x,y,S_0)\imply \neg on(y,x,S_0) )\land \\
\forall x\forall y \forall z( on(y,x,S_0)\land on(z,x,S_0)\imply y\!=\!z )\land \\
\forall x\forall y\forall z( on(x,y,S_0)\land on(x,z,S_0)\imply y\!=\!z ).
\end{array}
$\\
Notice we did not include any state constraint (axioms uniform in $s$) into BAT,
e.g., 
$\forall x\forall y\forall z\forall s.\ on(y,x,s)\land on(z,x,s)\imply y\!=\!z$. 
As stated in \cite{Reiter2001}, they are entailed from the similar sentences 
about $S_0$ for any situation that includes only consecutively possible actions.

  Finally, the foundational axioms $\Sigma$ are generalization of axioms for a
single successor function (see Section 3.1 in \cite{Enderton}) since SC has
a family of successor functions $do(\cdot,s)$, and each situation may have
multiple successors. As argued in \cite{Enderton}, the complete FO theory of 
single successor has countably many axioms, but it has non-standard models.
To eliminate undesirable non-standard models for situations, 
by analogy with Peano second-order (SO) axioms for non-negative integers,
where the number $0$ is similar to $S_0$, \cite{Reiter1993} proposed 
the following axioms for situations (with implicit $\forall s_1,a_1,s_2,a_2,s,a,s'$).
\begin{description}
\item
\hspace*{-2mm}	$
	do(a_1,s_1)\!=\!do(a_2,s_2) \imply a_1\!=\! a_2\land s_1\!=\! s_2$
\item
\hspace*{-2mm} 	$\neg(s\sqsubset S_0)$
\vspace{-1mm}
\item
\hspace*{-2mm}	$
	s \sqsubset do(a,s') \liff s\sqsubseteq s'$,
where $s\sqsubseteq s' \isdef \ (s\sqsubset s' \lor s\!=\!s')$
\item	
\hspace*{-2mm}
$\forall P.\ \big(P(S_0)\land \forall a\forall s(P(s)\imply P(do(a,s)) )\, \big)
	\imply \forall s(P(s))$
\end{description}
The last second order axiom limits the sort \textit{situation} to the smallest 
set containing $S_0$ that is closed under the application of 
$do$ to an action and a situation.
These axioms say that the set of situations is really a tree: there are
no cycles, no merging. These foundational axioms $\Sigma$ are domain independent.
Since situations are finite sequences of actions, they can be implemented
as lists in PROLOG, e.g., $S_0$ is like [~], and $do(A,S)$ adds an action $A$
to the front of a list representing $S$, i.e. $[A | S]$.
Therefore, in PROLOG, all situation terms implicitly satisfy the above
foundational axioms \cite{Reiter2001} and no SO reasoning is required. 

  It is often convenient to consider only executable (legal) 
situations: these are action histories in which it is actually possible 
to perform the actions one after the other.\\
$s < s' \isdef s\! \sqsubset s'\! \lland 
\forall a\forall s^* (\!s \sqsubset do(a,s^*\!) \sqsubseteq s'\imply Poss(a,s^*))$\\
where $s < s'$ means that $s$ is an initial sub-sequence of $s'$ and all
intermediate actions are possible.
Subsequently, we use the following abbreviations: \ 
    $s \leq s'\isdef (s < s') \lor s\!=\!s'$.\ 
Also, $executable(s)\isdef S_0 \leq s.$ \cite{Reiter2001} formulates
\begin{theorem}\label{executability} 
$executable(do([\alpha_1,\cdots,\alpha_n],S_0)) \liff$\\
\hspace*{2mm}
$poss(\alpha_1,S_0)\land \bigwedge_{i=2}^{n}poss(\alpha_i, do([\alpha_1,\cdots,\alpha_{i-1}],S_0)).$
\end{theorem}
\begin{theorem}\label{relative-satisfiability}
\cite{PirriReiterACM1999}
A basic action theory 
${\cal D}\!=\!\Sigma \land {\cal D}_{ss} \land {\cal D}_{ap} \land {\cal D}_{una}\land {\cal D}_{S_0}$ 
is satisfiable \ iff \ ${\cal D}_{una}\land {\cal D}_{S_0}$ is 
satisfiable.
\end{theorem}
Theorem~\ref{relative-satisfiability} states that no  SO axioms $\Sigma$ are 
needed to check for satisfiability of BAT $\bat$. This result is the key to 
tractability of $\bat$, since ${\cal D}_{una}\land {\cal D}_{S_0}$ are sentences in FOL.

  There are two main reasoning mechanisms in SC. 
One of them relies on the regression operator \cite{Waldinger77,Reiter1991} that
reduces reasoning about a query formula uniform in a given situation $\sigma$
to reasoning about regression of the formula wrt $\dsz$. Another mechanism 
called progression \cite{LinReiter1997} is responsible for reasoning forward,
where after each action $\alpha$ the initial theory $\dsz$ is updated to a
new theory $\dsa$. In this paper, we focus on progression in a local effect BAT, 
where SSAs are essentially quantifier free, as defined before. For a local effect
BAT, \cite{LiuLakemeyer2009} show that in a general case the size of $\dsa$ can 
increase in comparison to the size of $\dsz$. For this reason, the paper 
\cite{LiuLakemeyer2009} considers a special case of a $proper^+$ $\dsz$. 
The $proper^+$ theories are proposed in \cite{LakemeyerLevesqueKR2002} as a 
generalization of a $proper$ KB \cite{Levesque1998}, which is equivalent to 
a possibly infinite consistent set of ground literals. 
Let $e$ be an {\it ewff}, a well-formed formula whose only predicate is equality, 
and let a clause $d$ be a disjunction of fluent literals. Then, the universal 
closure $\forall(e \supset d)$ is called a {\it guarded clause}, or a 
$proper^+$-formula. An initial theory is in a $proper^+$ form, if it is 
a finite non-empty set of guarded clauses supplemented with the axioms of 
equality and the set of UNA for constants. The $proper^+$ initial theories 
generalize databases by allowing incomplete disjunctive knowledge about some of
the named elements of the domain \cite{LakemeyerLevesqueKR2002}. It turns out,
that for a local effect BAT with essentially quantifier free contexts,
if an initial theory $\dsa$ is in $proper^{+}$ form, then its progression $\dsa$
can be computed efficiently, namely in linear time wrt the size of $\dsz$
\cite{LiuLakemeyer2009}. Therefore, progression can be consecutively
computed for arbitrarily long sequences of actions. 

The \textit{Domain Closure Assumption} (DCA) for objects 
\cite{Reiter1977,DBLP:journals/jacm/Reiter80} 
means that the domain of interest is finite, the names of all objects in $\dsz$ 
are explicitly given as a set of constants $C_1, C_2,\ldots,C_K$, 
and for any object variable $x$ it holds that $\forall x (x\eq C_1 \lor x\eq C_2\lor \ldots\lor x\eq C_K)$
 According to the \textit{Closed World Assumption} (CWA), an initial theory ${\cal D}_{S_0}$ 
is conjunction of ground fluents, and all fluents not mentioned in ${\cal D}_{S_0}$ 
are assumed by default to be false \cite{Reiter1978,DBLP:journals/jacm/Reiter80}. 
According to an opposite, \textit{Open World Assumption} (OWA), an initial 
theory $\dsz$ can have a more general form, e.g., it can be in a $proper^+$ form. 
As proved in
Theorem 4.1 in \cite{DBLP:journals/jacm/Reiter80}, in the case of a data base 
augmented with the axioms of equality, the queries that include only 
$\exists$-quantifiers over object variables can be answered without DCA.
Similar results can be proved for a  $\dsz$  in a $proper^+$ form, 
assuming there are no object function symbols other than constants.
From this fact, the above mentioned results, and the results from \cite{YongmeiPhD2005},
it follows that in the case of a BAT where $\dsz$ is in a  $proper^+$ form, 
the context conditions in SSAs are essentially quantifier free,
where the preconditions $\Pi_A(\bar{x},s)$ in $\mathbf{{\cal D}_{ap}}$ 
include only $\exists$-quantifiers over object variables, the goal formula 
includes only $\exists$-quantifiers over object variables, and all sets of
axioms use only a bounded number of variables, the length-bounded planning problem 
can be solved without DCA over the object variables (and without CWA). 
In the next section, we formulate the (bounded) planning problem for BATs
and show a planner can be developed from the first principles.

\section{Bounded Lifted Planning with BATs} 
Let $G(s)$ be a goal formula that is  uniform in $s$ and has no other free variables.
Let $Length(s)$ be a number of actions in situation $s$, i.e., 
 $Length(do([\alpha_1,\cdots,\alpha_N],S_0) )\eq N$ and $Length(S_0)\eq 0$.
Following \cite{Reiter2001}, the bounded planning problem can be formulated
in SC as
\begin{equation}\label{planningReiter}
\bat \models \exists s.\ Length(s) \leq N \land executable(s) \land G(s),
\end{equation}
where $N\geq 0$ is an upper bound. From the Theorem~\ref{executability}, 
definition of $executable(s)$ and the foundational axioms $\Sigma$, it follows
this can be equivalently reformulated for $N> 2$ as\\
$
\begin{array}{c}
\bat\models G(S_0)\lor \exists a_1 \big(poss(a_1,S_0)\land G( do(a_1,S_0) )\big) \lor \\
    \exists a_1\exists a_2 \big(\, S_0 < do([a_1,a_2],S_0)\lland\ \big(\, G(do([a_1,a_2],S_0))\ \lor \\
    \exists s (do([a_1,a_2],S_0) \leq s \land Length(s)\lleq N \land G(s))\, \big)\, \big).
\end{array}
$\\
This simply means that if there exists  a situation term that solves 
the planning problem (\ref{planningReiter}), then either it is $S_0$, or for 
some action $a_1$ that is possible in $S_0$, it is $do(a_1,S_0)$, or for some
actions $a_1$ and $a_2$ that are consecutively possible from $S_0$, either
$G(do([a_1,a_2],S_0))$ holds, or there exists situation $s$ that is
executable from $do([a_1,a_2],S_0)$ such that its total length is less than or
equal to $N$ and the formula $G(s)$ holds in $s$. Suppose that a BAT $\bat$ has $k$
different action functions $A_1(\bar{x_1}),\ldots,A_k(\bar{x_k})$. Then, 
according to the domain closure axioms for actions, any formulas $\exists a\, \psi(do(a,S_0) )$ 
and $\exists a\exists a'\,  \psi( do(a',do(a,S_0)) )$ are respectively equivalent 
to $\bigvee_{\!i\eq 1}^k \exists \bar{x_i} \psi\big( do(A_i(\bar{x_i}), S_0)\big)$ and 
$\bigvee_{\!j\eq 1}^k \! \exists \bar{x_j}\! \bigvee_{\!i\eq 1}^k \!\exists \bar{x_i} \psi\big(\!do(A_j(\bar{x_j}),\! do(A_i(\bar{x_i}), S_0) ) \big)$.
Therefore:
\begin{theorem}\label{planningExpanded}
A ground situation term $do([\alpha_1,\!\cdots\!,\alpha_n],S_0)$, $n\lleq N$ is 
a solution to problem (\ref{planningReiter})  iff 
for some sequence $(i_1,\cdots,i_n)$ of action indices, $1\leq i_j\leq k$,
there are ground substitutions
for action arguments that unify $A_{i_1}(\bar{x_{i_1}})$ with $\alpha_1$,\ldots,
$A_{i_n}(\bar{x_{i_n}})$ with $\alpha_n$, and for these substitutions both\\
$\exists \bar{x_{i_n}} \cdots \exists \bar{x_{i_1}}
     G\big( do([A_{i_1}(\bar{x_{i_1}}),\!\cdots\!,A_{i_n}(\bar{x_{i_n}})], S_0) \big)$
and the formula 
$\exists \bar{x_{i_n}}\! \cdots\! \exists \bar{x_{i_1}}
     S_0\!\lleq do([A_{i_1}\!(\bar{x_{i_1}}),\!\cdots\!,A_{i_n}\!(\bar{x_{i_n}})],\! S_0)$
are entailed from a BAT $\bat$.
\commentout{ 
    $\bigvee_{\!i_n\eq 1}^k \! \exists \bar{x_{i_n}}\! \cdots \bigvee_{\!i_1\eq 1}^k \!\exists \bar{x_{i_1}}
         G\big(\!do[(A_{i_1}(\bar{x_{i_1}}),\!\cdots\!,A_{i_n}(\bar{x_{i_n}})], S_0) \big)$\\
    and the formula \\
    $\bigvee_{\!i_n\eq 1}^k \! \exists \bar{x_{i_n}}\! \cdots\!\! \bigvee_{\!i_1\eq 1}^k \!\exists \bar{x_{i_1}}
         S_0\!\lleq do([A_{i_1}\!(\bar{x_{i_1}}),\!\cdots\!,A_{i_n}\!(\bar{x_{i_n}})],\! S_0)$
    are entailed from a BAT $\bat$.
}   
\end{theorem}
This theorem is the first key observation that helps design a lifted planner 
based on SC. The planner has to search over executable sequences of actions 
on a situation tree. Note that the state space and states themselves remain implicit, 
since situations serve as symbolic proxies to states. Whenever 
a sequence of $i$ ground actions determined by search results in a situation 
    $do([\alpha_1,\cdots,\alpha_i],S_0))$,
to find the next action the planner must check among the actions
$A_1(\bar{x_1}),\ldots,A_k(\bar{x_k})$ for which of the values of their object 
arguments these actions are possible in $do([\alpha_1,\cdots,\alpha_i],S_0))$.
Since this computation is done at run-time, but not before the planner starts 
searching for actions, the SC-based planner is naturally lifted, 
no extra efforts are required.

The second key observation is that an efficient deductive planner needs control 
that helps select for each situation the most promising next possible action 
to execute. This control can be provided by an A$^*$ algorithm that relies on 
a domain independent heuristic function. 
In the next section, we show how our heuristic deductive planner can be implemented
in PROLOG. This implementation requires both DCA and CWA. A more general implementation
is left to future work. 

\section{Implementation}
Our SC-based TPLH planner is implemented in PROLOG following the two 
key observations mentioned in the previous section. The planner is driven by 
theorem proving that is controlled by a version of A$^*$ search for a shortest
sequence of actions that satisfies  (\ref{planningReiter}).  The distinctive 
feature of TPLH is that it does forward search over the situation tree from $S_0$.
Since each search node is a unique situation, the previously visited nodes
cannot be reached by search again, and there is no need to keep them in memory.
Moreover, frontier nodes cannot be reached 
along different paths since each situation represents a unique path on the situation
tree. Therefore, there is no need to check whether new neighbors to be 
explored have been already included in the frontier. For simplicity, the cost 
of every action is considered to be 1, and the cost of a path to a node is 
simply the length of situation representing the node.  This search
terminates as soon as it finds a ground situation $S$ that satisfies a goal $G(S)$.
In Algorithm \ref{alg1}, a plan is a situation that is represented as a list of 
actions from $S_0$, while $S_0$ is represented as the empty list.

\begin{algorithm}[t]
\caption{$A^*$ search over situation tree to find a plan}\label{alg1}
\algorithmicrequire  $(\bat,G)$ - a BAT $\bat$ and a goal formula G \\
\algorithmicrequire $H$ - Heuristic function \\
\algorithmicrequire $N$ - Upper-bound on plan length \\
\algorithmicensure  $S$ that satisfies (\ref{planningReiter}) \Comment{Plan is the list of actions in $S$}

\begin{algorithmic}[1] 
    \Procedure{Plan}{$\bat,G,N,H,S$}
        \State $PriorityQueue \gets \varnothing$ \Comment{Initialize PQ}
        \State $S_{0}.Val \gets (N+1)$
        \State $PriorityQueue.insert(S_0, S_{0}.Val)$
        \State $Init \gets \text{InitialState}(\dsz)$ \Comment{Initialize state}
        \While{$ \textbf{not } PriorityQueue.empty()$}
        \State $S \gets PriorityQueue.remove()$   
        \State $Now \gets \text{\ Progress}(Init,S)$ \Comment{Current state}
            \If{\text{\ Satisfy($Now,G)\ $}}
                \State \textbf{return} $S$ \Comment{Found list of actions}
            \EndIf
            \State $Acts \gets \text{FindAllPossibleActions}(Now)$
            \If{$Acts == \varnothing$}
                         \State \textbf{continue} \Comment{No actions are possible in $S$}
            \EndIf
            \For{$A_{i} \in Acts$}
                \State $S_n \gets do(A_i,S)$ \Comment{$S_n$ is next situation}
                \State $St \gets \text{\ Progress}(Now,A_i)$ \Comment{Next state}
                    \If{$\ \text{Length}(S) \geq N$}
                         \State \textbf{continue} \Comment{$S_n$ exceeds upper bound}
                    \Else \ $d \gets N\!-\!\text{Length}(S)$ \Comment{$d$ is depth bound}
                     \EndIf
                \State \hspace*{-1mm}$S_n.Val \gets \text{Length}(S_n)\! +\! H(\bat,\!G,\!d,\!S_n,\!St)$
                \State $PriorityQueue.insert(S_n,S_n\mathbf{.}Val)$
            \EndFor
        \EndWhile
        \State \textbf{return} $False$\Comment{No plan for bound $N$}
    \EndProcedure
\end{algorithmic}

\end{algorithm}

The main advantage of this design is that the frontier stored in a priority
queue consists of situations and their $f$-values\footnote{This is a term  from 
the area of heuristic search, see \cite{GeffnerBonetBook2013,GhallabNauTraverso2004}. 
There are plan costs $g(s)$ (the number of actions in $s$), and there are heuristic 
estimates ($h$ values) of the number of actions remaining before the goal can be 
reached. The total priority of each search node (in our case it is a situation 
$s$) is estimated as $f(s)\!=\!g(s)\!+\!h(s)$. A smaller total effort $f(s)$ indicates a more 
promising successor situation $s$.}
 computed as the sum of situation length and a heuristic estimate. 
Therefore, situations serve
as convenient symbolic proxies for states. As usual, a state corresponding to 
situation $s$ is a set of fluents that are true in $s$. 
However, in hard-to-ground domains, each state can be very large, 
and storing all intermediate states can exhaust all memory. 
This issue was demonstrated on realistic domains such as Organic Synthesis 
\cite{MasoumiAntoniazziSoutchanski2015}.  Moreover, in the case of planning in 
physical space and real time, the state space is infinite, but a deductive 
planner can still search (without ad-hoc 
discretizations of space and time) over finite sequences of actions  according 
to semantics in \cite{BatusovSoutchanskiICAPS2019}.

In Algorithm \ref{alg1}, the sub-procedure InitialState($\dsz$) on Line 5
takes the initial theory as its input, and computes the initial state
under the usual DCA and CWA. (Note this is a limitation of the current 
implementation, but not of the TPLH approach in general.)
We store this initial state $Init$ in a specialized data structure 
that facilitates computing progression efficiently.
On the Line 7, the algorithm extracts
the next most promising situation $S$ from the frontier. Then, on Line 8, it 
computes progression $Now$ of the initial state using the actions mentioned in $S$.
This is rather straightforward update of the initial state $Init$ that is facilitated
by our data structure. Note that in the case of incomplete $\dsz$,
and a local effect BAT, one would need a more sophisticated algorithm for 
progression, as explained before. On Line 9, there is a check whether the goal 
formula $G$ is satisfied in the current state $Now$. If it is, then $S$ is 
returned as a plan. If not, then on Line 12, the algorithm finds all actions
that are possible from the current state using the precondition axioms. 
In fact, the sub-procedure $FindAllPossibleActions$ is using preconditions
to ground all action functions from the given BAT in the current state. 
Since actions are grounded at run-time, TPLH is a lifted planner by design.
If there are no actions possible from $N\!ow$, then the algorithm proceeds to
the next situation from the frontier. Otherwise, for each possible ground
action $A_i$, it constructs the next situation $S_n\eq do(A_i,S)$, and if its
length does not exceed the upper bound $N$, it computes the positive integer 
number $d$ on Line 21 as $N\!-\!Length(S)$. This bound $d$ is provided as an input 
to the heuristic function $H(\bat,G,d,S_n,St)$ that does limited look-ahead
up to depth $d$ from $St$ to evaluate situation $S_n$. On Line 24, $S_n$ and
its $f$-value  $S_n.Val$ are inserted into the frontier, and then search
continues until the algorithm finds a plan, or it explores all situations
with at most $N$ actions. The \textbf{for}-loop, Lines 16-24, makes sure that
all possible successors of $S$ are constructed, evaluated and inserted into 
the frontier. This is important to guarantee completeness of Algorithm~\ref{alg1}.
The bound $N\geq 0$ makes sure that search will always terminate in a finite
domain, since there are finitely many ground situations with length less than or
equal to $N$, and in the worst case, all of them will be explored.  However, 
due to this upper bound, search may terminate prematurely, i.e., without 
reaching a goal state, if the shortest plan includes more than $N$ action. 
Consequently, this planner is complete only if the bound $N$ is greater than 
or equal to the length of a shortest plan. 
Obviously, the planner is sound. 

Note that $Progress(Init,S)$ computes afresh the current state 
from the given initial state and the list of actions in $S$.
If the computed state $N\!ow$ does not satisfy a goal formula, it is not preserved
after computing heuristic value from $N\!ow$. Only successor situations are retained 
in the frontier, but not their corresponding states. This is an important
contribution of TPLH approach. The previous planning algorithms usually retained 
states, but not situations in their frontiers: see Discussion section for details.
Only the initial  state $Init$ remains in memory, but all other intermediate
states are recomputed from $Init$ when needed. Therefore, TPLH trades speed for 
memory. Since the memory footprint of TPLH is smaller than it is for alternative
implementations, our approach is suitable for planning in hard to ground 
domains. 

\begin{algorithm}[tb]
	\caption{GraphPlan heuristic with delete relaxation}\label{alg2}
	\algorithmicrequire $(\bat, G)$ - BAT $\bat$ and a goal formula $G$ \\
	\algorithmicrequire  $d\geq 1$ - Look-ahead bound for the heuristic algorithm \\
	\algorithmicrequire $S_n,L$ - The current situation and its length \\
	\algorithmicrequire $St$ - The current state \\
	\algorithmicensure  {\it Score} - A heuristic estimate for the given situation
	
	\begin{algorithmic}[1] 
		\Procedure{$H$}{$\bat,G,d,S_n,St$}
		\State $Depth \gets 0$ 
		\State $PG \gets \langle S_n,St\rangle$ \Comment{Initialize Planning Graph}
		\While{$\textbf{not} \: \textrm{Satisfy}(St,G)\ \textbf{and\ } Depth\leq d$}
		\State $\{ ActSet\} \gets \text{FindAllPossibleActions}(St)$
		\State $N\!ew\!Acts\! \gets \textrm{Select relevant actions from} \ ActSet$
		\State $St \gets \textrm{ProgressRelaxed}(St,N\!ew\!Acts)$ \\
		     \Comment{Add all new positive effects $N\!ew\!E\!f\!f\!s$ to the state}
		\State $N\!ext\!Layer \gets \langle N\!ew\!E\!f\!f\!s,N\!ew\!Acts,St \rangle$\\
		 \Comment{Record actions added, their effects, the current state}
		\State $PG.extend(N\!ext\!Layer)$
		\State $Depth \gets Depth+1$ 
		\EndWhile
        \State $Goal \gets $ \textrm{Convert $G$ into a set of literals}  
         \If{$Depth > d$} \textbf{return} $(L+d)$   \Comment{Penalty}
         \Else\ \ \textbf{return} $Reachability(\bat,Goal,PG)$
                     \EndIf
		\EndProcedure
	\end{algorithmic}
	
\end{algorithm}

\begin{algorithm}[h]
	\caption{Reachability score for a set of goal literals}\label{alg3}
	\algorithmicrequire  $(\bat,G)$ - A BAT $\bat$ and a set $G$ of goal literals \\
	\algorithmicrequire $PG$ - A planning graph, initialized to $\langle S_n,St\rangle$ \\
	\algorithmicensure  $V$ - A heuristic estimate for achieving $G$
	
	\begin{algorithmic}[1]
		\Procedure{$Reachability$}{$\bat,G,PG$}
		\If{\ $PG==\langle S_n,St\rangle$}  \ 	\textbf{return} 0
		\Else \, $\langle Ef\!f\!s,\! Acts,\! St \rangle\! \gets PG.removeOuterLayer$
		\EndIf
		\State \hspace*{-3mm}
$CurrGoals\! \gets\! G \cap E\!f\!f\!s$ \, \Comment{The set of achieved goals}
		\State \hspace*{-1mm}
$NewGoals \gets \emptyset$      \Comment{To collect preconditions}
		\State \hspace*{-1mm}
$BestSupport \gets \emptyset$      \Comment{Easiest causes for CurrGoals}
		\For{$g \in CurrGoals$}
		\State \hspace*{-3mm}
		 $Relev\! \gets$ \{actions from $Acts$ with $g$ as add effect\}
		\For{\ $a \in Relev$\ }
		\State $a.Pre \gets $ \{the set of preconditions of $a$\}
		\State $a.Estimate\! \gets Reachability(\bat,\!Pre,\!PG)$
		\EndFor
		\State $Best\!Act \gets$ \textrm{ArgMin} \{$a.Estimate$ over $Relev$\}\\
		 \Comment{Find the action from $Relev$ with minimum estimate} 
		\State $NewGoals \gets NewGoals \cup Best\!Act.Pre$
		\State $BestSupport \gets BestSupport \cup Best\!Act$
		\EndFor
		\State $RemainGoals \gets G - CurrGoals$
		\State $NextGoals \gets RemainGoals \cup NewGoals$
		\State $C_1 \gets \textrm{Count}(BestSupport)$ \Comment{i.e. \# of best actions}
		\State $C_2 \gets Reachability(\bat,NextGoals,PG)$
		\State \textbf{return} $C_1 + C_2$
		\EndProcedure
	\end{algorithmic}
\end{algorithm}

Computing the heuristic function is done in two stages, with usual delete relaxation:
Algorithm \ref{alg2}. First, a planning graph is built from the current state, 
layer-by-layer until all goal literals are satisfied. Supporting actions are then 
found for the goal literals, going backwards through the graph.

The planning graph is initialized to the current situation and state. 
At each step in building the planning graph, all possible actions for 
the current state are found, and then filtered so that only those actions 
with one or more new (positive) add effects not in the current state are kept. 
The state is updated using relaxed progression  to incorporate their new add 
effects. These `relevant actions', their new add effects and the updated state are 
inserted into the next layer of the planning graph, and the process is repeated.

Once all goal literals are satisfied, the most recent layer of the planning 
graph is examined. For each of the new add effects in this layer belonging to 
the set of goal literals, all relevant actions from the layer which achieve 
the effect are selected. These are referred to as the `supporting actions' 
for the goal literal. 
For each supporting action, a `reachability' score is recursively computed using 
its preconditions as the new goal literals. The easiest action whose preconditions 
have the lowest reachability is considered the `best supporting action'. Thus,
our reachability score represents the estimated cost of achieving a set of literals. 
If all literals are satisfied in the initial layer of $PG$, 
then the set's reachability is 0. Otherwise, its reachability is equal to 
the reachability of the remaining goals and preconditions for the set of  
the easiest actions, plus the number of best support actions. 

\section*{Experimental Results}
To evaluate our implementation experimentally, we run our planner on several 
STRIPS benchmarks, where preconditions of actions are conjunctions of 
fluents (though they can include negations of equality between variables or 
constants), the SSAs have no context conditions, and the goal formula is 
a conjunction of ground fluents. 

Tests were run separately using the TPLH, FD, and BFWS planners. 
TPLH and FD used the A$^*$ algorithm to prioritize shorter plan lengths, whereas 
BFWS used a default greedy search algorithm. Both TPLH and FD did eager search 
with FF heuristic. All testing was done on a desktop with an Intel(R) Core(TM) 
i7-3770 CPU running at 3.40GHz. 
Tests measured total time spent, plan length, 
and number of states (situations) visited. Comparisons are made based on 
the average ratio of the values for TPLH to those of FD and BFWS across all 
problems in a given domain. The TPLH planner, domain files and problem instances
have been loaded, compiled and run within ECLiPSe Constraint Logic Programming
System, Version 7.0 \#63 (x86\_64\_linux), released on April 24, 2022.
In comparison, the FD and BFWS were compiled into executable files.
This difference should be taken into account while reviewing
the results below.
For example, as shown in Table \ref{TimeVisitedLength}, on BW, TPLH spent 
an order of $10^2$ more time per plan step than FD, and an order of $10^4$ 
more time per plan step than BFWS. Recall that BFWS does default greedy 
best first search, and 
this is why it takes less time than FD that does A$^*$ eager search 
with FF heuristic.

\subsection*{Domains and Problem Generation}

Testing was done over randomly generated problems for 8 different popular domains
that represent well the variety of planning problems from the competitions. These domains were Barman, BlocksWorld, ChildSnack, Depot, FreeCell, Grippers, Logistics, and Miconic. In addition, testing was also done on 10 pre-existing problems belonging to the PipesWorld domain. All domains are in STRIPS, extended to include negated equalities and object typing. For simplicity, the Barman domain was modified to remove action costs.

Roughly 100 problems with varying numbers of objects were generated for each of the specified domains, 
using publicly available PDDL generators. All PDDL domains and generated 
instances files were automatically translated from PDDL to PROLOG using our 
program that constructs a hash table based representation of an initial theory.
The TPLH planner was run over every problem using a 30min time-out limit, 
and a 50 MB global stack size limit. Problems for which the planner 
timed out were discarded, as were problems with 0-step solutions 
(i.e., where the initial state satisfied the goal state). 
The number of kept instances for each domain is shown in parentheses 
after the domain name in Table~\ref{TimeVisitedLength}.
The TPLH planner was given the upper bound $N\!=\!100$ for all planning
instances that usually had short solutions, e.g., 20 steps or less. Recall $N$ 
is used to guarantee completeness of TPLH, but it has little effect in this set 
of experiments. Namely, when we tried different values $N\!=\!\{25,50,75,100,125,150\}$
over some domains, the total time varied within 0.5\%, and plan length and 
the number of situations visited by TPLH did not change at all.

Before TPLH could be tested on a domain, the domain file was converted from PDDL to PROLOG, and initial state hash tables were built for each individual problem. Translating domains files themselves took very little time (under 0.1 seconds in all cases), and this cost was further amortized by the fact that it only needed to be done once, regardless of how many problems were tested. Building initial state hash tables however could take a non-negligible amount of time. This time was consistent across all problems belonging to a domain, and ranged from approximately 1.5 seconds per problem (for BlocksWorld) to nearly 10 seconds per problem (for Barman). The inefficiency here is tied to the current implementation of the script used, and is not inherent to the task of creating the hash tables themselves. Preprocessing time for each problem was added to the time spent by the planner itself to get the total time required to solve a problem. (Performance of TPLH on easy instances was much better when preprocessing was factored out.)

\commentout{    
\begin{table}[b]
	\centering
	\begin{tabular}{|l|c|c|c|}
		\hline
		{} & \textbf{Tested} & \textbf{Kept} & \textbf{Ratio} \\ \hline
		\textbf{Barman} & 100 & 100 & 1 \\ \hline
		\textbf{BlocksWorld} & 100 & 95 & 0.95 \\ \hline
		\textbf{ChildSnack} & 100 & 100 & 1 \\ \hline
		\textbf{Depot} & 105 & 76 & 0.72 \\ \hline
		\textbf{FreeCell} & 120 & 95 & 0.79 \\ \hline
		\textbf{Grippers} & 100 & 98 & 0.98 \\ \hline
		\textbf{Logistics} & 134 & 119 & 0.89 \\ \hline
		\textbf{Miconic} & 100 & 93 & 0.93 \\ \hline
		\textbf{PipesWorld} & 10 & 6 & 0.6 \\ \hline
	\end{tabular}
	\caption{The number of problems tested for each domain, and the fraction of which were included.}
	\label{pddlProblems}
\end{table}
}

\subsection*{Plan Lengths and Number of Situations Visited}

When testing the problems using TPLH, the number of situations visited was recorded, as was the length of the produced plan and the total time taken. A situation was considered as having been visited upon checking whether it satisfied a goal state. Thus, the minimum number of \textit{situations} visited by TPLH is one greater than the length of the produced plan. The same data was gathered when testing using the FD and BFWS planners, with the distinction that the number of \textit{states} visited by each planner was recorded, rather than situations. First, we compare TPLH to FD, and second, 
we compare TPLH to BFWS, see Table~\ref{TimeVisitedLength}.

\begin{table}[!b]
	\centering
	\begin{tabular}{|l| c  c  c | c  c  c |}
		\hline
		{} &  
		\multicolumn{3}{c|}{\textbf{FD}} & \multicolumn{3}{c|}{\textbf{BFWS}} \\ \hline
		\textbf{\small Barman} (100) & \!1  & 0.10 & $10^2$\! & 0.90  & 0.24 & $10^4$\!\\ \hline
		\textbf{\small BW} (95) & \!0.98 & 0.11 & $10^2$\!& 0.62 & 0.56 & $10^3$\!\\ \hline
		\textbf{\small ChldSnk}(100)\!& 1 & 0.09 & $10^2$\!& 0.96 & 23.75 & $10^4$\!\\ \hline
		\textbf{\small Depot} (76) & 1.00 & 0.09 & $10^2$\!& 0.98 & 2.81 & $10^4$\!\\ \hline
		\textbf{\small FreeCell} (95) & 1.01 & 0.10 & $10^2$\!& 0.99 & 0.23 & $10^4$\!\\ \hline
		\textbf{\small Grippers} (98) & 1.00 & 0.06 & $10^2$\!& 0.79 & 6.39 & $10^3$\!\\ \hline
		\textbf{\small Logistcs}(119) & 1 & 0.17 & $10^3$\!& 0.87 & 4.07 & $10^4$\!\\ \hline
		\textbf{\small Miconic} (93) & 1 & 0.11 & $10^2$\!& 0.86 & 0.26 & $10^3$\!\\ \hline
		\textbf{\small PipesWrld}(6) & 1 & 6.06 & $10^2$\!& 0.94 & 15.27 &$10^3$\!\\ \hline
	\end{tabular}
	\caption{The average ratio of plan length, situations/states visited, and time per step for TPLH plans to those produced by FD, BFWS. The lower numbers indicate better performance of TPLH.}
	\label{TimeVisitedLength}
\end{table}

The number of situations visited by TPLH is considerably smaller than the number of states visited by FD across eight of the nine domains that were tested: see the middle column in Table~\ref{TimeVisitedLength}. In this regard, TPLH outperformed FD on all randomly generated problems except for five (one of these belonging to the BlocksWorld domain, and the other four to Logistics). TPLH also managed to produce plans that were mostly on-par with FD in terms of plan length, see the left most column in  Table~\ref{TimeVisitedLength}. In five of the domains, each planner produced plans of exactly equal length across all problems. Out of the other four domains, FD had a slight edge in three of them. 
The PipesWorld domain is a notable outlier from the others in that TPLH had to visit far more situations on average than both FD and BFWS.

When comparing to BFWS, TPLH produced shorter average plans across all nine domains, 
see the left column related to BFWS in Table~~\ref{TimeVisitedLength}. 
This is not surprising since BFWS does greedy search, but TPLH does A$^*$ search. 
When comparing the average number of situations/states visited, TPLH outperformed 
BFWS in four of the domains, but was beaten in the ChildSnack, Depot, Grippers, 
Logistics, and PipesWorld domains, see the middle column in  Table~\ref{TimeVisitedLength}. 
Looking into the data, it appears that for three of the domains, Depot, Grippers, 
and Logistics, TPLH beat BFWS for a majority of the problems, but lost overall. 
In other words, TPLH actually visited fewer situations in the majority of the 
problems; in the ones which it lost, however, it did so by a large margin, see 
the right-most column in  Table~\ref{shorterPlansFewSits}. Additionally, TPLH beat BFWS in 
this regard on exactly 50\% of the problems from the PipesWorld domain.

In Table~\ref{TimeVisitedLength}, the middle column related to BFWS, you see TPLH 
visited on average 23.75 times more situations than  the number of states visited by BFWS.
A few inherent aspects of the ChildSnack domain lead to the heuristic performing 
poorly wrt BFWS. Firstly, plans in this domain are highly 'interleavable'; i.e. 
there are several permutations of the same actions which are all valid solutions. 
Secondly, ChildSnack problems have relatively few goal atoms, which are all 
achieved by the last few actions of a plan. Third, no heuristic is perfect. 
As the heuristic is domain-independent, it is natural that there will be some 
domains where it excels, and some where it struggles. Notice that TPLH performs 
better than FD with a similar FF heuristic in terms of the ratio situations/states 
visited: our ratio is 0.09 of FD. 

Refer to Table  \ref{shorterPlansFewSits}  
for \% of problems across each domain for which TPLH performed at least 
as well as its competitors on plan length (left column) and situations visited
(right). 

\begin{table}
	\begin{tabular}{|l|c c|c c|}
		\hline
		{} & \multicolumn{2}{c|}{\textbf{FD}} & \multicolumn{2}{c|}{\textbf{BFWS}} \\ \hline
		\textbf{Barman} & 1 & 1 & 1 & 0.97 \\ \hline
		\textbf{BlocksWorld} & 0.92 & 0.99 & 1 & 0.96 \\ \hline
		\textbf{ChildSnack} & 1 & 1 & 1 & 0.40 \\ \hline
		\textbf{Depot} & 0.99 & 1 & 0.99 & 0.78 \\ \hline
		\textbf{FreeCell} & 0.91 & 1 & 0.92 & 1 \\ \hline
		\textbf{Grippers} & 0.96 & 1 & 1 & 0.59 \\ \hline
		\textbf{Logistics} & 1 & 0.97 & 1 & 0.69 \\ \hline
		\textbf{Miconic} & 1 & 1 & 1 & 0.98 \\ \hline
		\textbf{PipesWorld} & 1 & 0.5 & 1 & 0.5 \\ \hline
			\end{tabular}
	\caption{Percentage of problems for each domain where TPLH produced a plan of shorter or equal length, and visited fewer situations than FD and BFWS.}
	\label{shorterPlansFewSits}
	\end{table}
%
\begin{table}
		\begin{tabular}{|l|c|c|}
		\hline
		{} & \textbf{Average r} & \textbf{\% s.t. $r \geq 0.75$} \\ \hline
		\textbf{Barman} & 0.29 & 0 \\ \hline
		\textbf{BlocksWorld} & 0.60 & 0.53 \\ \hline
		\textbf{ChildSnack} & 0.34 & 0.40 \\ \hline
		\textbf{Depot} & 0.57 & 0.51 \\ \hline
		\textbf{FreeCell} & 0.89 & 1 \\ \hline
		\textbf{Grippers} & 0.38 & 0.30 \\ \hline
		\textbf{Logistics} & 0.41 & 0.29 \\ \hline
		\textbf{Miconic} & 0.57 & 0.14 \\ \hline
		\textbf{PipesWorld} & 0.32 & 0.33 \\ \hline
				\end{tabular}
	\caption{Average ratio $r$ of plan length to number of situations visited by the TPLH planner, as well as the fraction of problems from each domain for which $r$ was greater than or equal to 0.75. }
	\label{hstcPerformance}
\end{table}

The heuristic used by TPLH performed remarkably well on certain problems, only ever visiting situations which were a subsequence of the final plan. In the FreeCell domain for example, this was true of every problem tested. This is likely due to the nature of the Planning Graph data structure and the process used for finding best supporting actions. When evaluating actions which achieve the goal state for the relaxed problem, the cost of achieving the preconditions of each action is recursively computed, and the action with the lowest such cost is selected. 

This means that for highly sequential problems, where a specific chain of actions is necessary to allow a sub-goal to be achieved (e.g. in FreeCell, cards must be placed on the foundation pile in sequential order), the heuristic can identify situations which allow for shorter causal chains. As long as a given move completes a step in this chain, TPLH recognizes the resulting situation as more promising than the previous one, and pursues it. When the causal chain is complete for the final goal, the problem is solved.

\commentout{    
\begin{table}[!b]
	\centering
	\begin{tabular}{|l|c|c|}
		\hline
		{} & \textbf{Average Ratio} & \textbf{Fraction s.t. $r \geq 0.75$} \\ \hline
		\textbf{Barman} & 0.29 & 0 \\ \hline
		\textbf{BlocksWorld} & 0.60 & 0.53 \\ \hline
		\textbf{ChildSnack} & 0.34 & 0.40 \\ \hline
		\textbf{Depot} & 0.57 & 0.51 \\ \hline
		\textbf{FreeCell} & 0.89 & 1 \\ \hline
		\textbf{Grippers} & 0.38 & 0.30 \\ \hline
		\textbf{Logistics} & 0.41 & 0.29 \\ \hline
		\textbf{Miconic} & 0.57 & 0.14 \\ \hline
		\textbf{PipesWorld} & 0.32 & 0.33 \\ \hline
	\end{tabular}
	\caption{Average ratio $r$ of plan length to number of situations visited by the TPLH planner, as well as the fraction of problems from each domain for which $r$ was greater than or equal to 0.75. }
	\label{hstcPerformance}
\end{table}
}   

Measuring the ratio $r$ of the length of the plan produced to the number of 
situations visited by TPLH, we can evaluate the performance of our heuristic 
across each of the nine domains tested. We used a cutoff value of $r \geq 0.75$
to identify the \% of problems that the heuristic guided effectively,
see Table~\ref{hstcPerformance}. 
As previously discussed, the heuristic was able to effectively
guide 100\% of problems in the FreeCell domain. It also performed well on the BlocksWorld domain (53\%) and the Depot
domain (51\%). 
At the lowest end,  
none of the problems from Barman met this threshold $r \geq 0.75$.

The recursive nature of finding the best supporting actions necessitates a lot of redundant computations. The current non-optimized implementation of the heuristic redoes these computations each time, and therefore spends the vast majority (upwards of 95 percent) of its time computing heuristic values. This is part of the reason why TPLH is orders of magnitude slower than FD and BFWS. 
Memoization would eliminate these repeated calculations, and has the potential to greatly increase the planner's efficiency.

\section{Discussion}
To the best of our knowledge, TPLH is the first deductive planner based on 
SC with a domain independent heuristic.
There were several earlier proposals to develop a situation calculus inspired
deductive planner in PROLOG  \cite{LinPlannerR,Reiter2001,LevesqueThinking2012}, 
but these planners do uninformed forward depth-first search, e.g., with an
iterative deepening strategy. They require both DCA and CWA.

  The earlier work on deductive planning in the 1980s-90s is well reviewed in
\cite{BiundoEWSP1993,Fronhofer1996,Fronhfer1997Inform,BibelAI1998}. 
The interested readers can find other references in those publications.
This earlier work includes well-known research on using the linear connection 
method that circumvents the frame problem and thereby facilitates deductive planning
\cite{BibelNGC1986,BibelAI1998,HolldoblerSchneebergerNGC1990}. 
Another line of research on deductive planning adapted linear logic, e.g., see
\cite{GrosseHolldoblerSchneeberger1996,CresswellSmaillRichardson1999},
and this research continues up to date.
However, to the best of our knowledge, the earlier work did not 
lead to a competitive implementation and did not use domain independent heuristics. 
\cite{Fronhofer1996} noted that to his surprise an 
implementation based on SC was comparable with an alternative implementations 
based on the linear connection method, and both deductive planners were 
competitive with the specialized planner UCPOP \cite{PenberthyWeldKR1992}. 
He compared run time using randomly generated small instances of BW and 
the briefcase planning problems. He concluded: 
``This opens up the question whether further exploitation of up-to-date theorem 
proving technology will bring about further gains in efficiency, and whether it
might lead eventually to a complete rehabilitation of Situational Calculus".

   There are several variants of conformant planning with an OWA. 
For example, a sound and complete conformant planner based on the situation calculus 
is described in the paper \cite{FinziPirriReiterOWA2000} and in
Chapter 10 of \cite{Reiter2001}. Their planner represents an initial theory
$\dsz$ as conjunction of prime implicates, it makes DCA and relies on regression 
of a goal formula, but it does depth-first forward search to compute a plan. 
The knowledge-level planner PKS described in \cite{PetrickPhD2006} works under OWA, 
and moreover, it exceeds the usual DCA since it allows the function symbols, 
e.g., the term $parentO\!f(john)$ can be an argument of an action. PKS focuses 
mostly on epistemic and contingent planning, it does search over state space, 
and for these reasons it is different in scope from TPLH. To the best of our 
knowledge, most publications in automated AI planning rely on DCA,
including previous work on conformant planners, e.g., see 
\cite{ToSonPontelliAI2015,HoffmannBrafmanAIJ2006}
that use representations different from situation calculus BATs, and they do not
consider lifted deductive planning. In the latter paper \cite{HoffmannBrafmanAIJ2006}, 
search goes over executable sequences of actions rather than state space, 
but there is no connection with SC, their system relies on a SAT solver, 
and therefore it can be problematic in the large, hard-to-ground domains.

However, since the progression of a $\dsz$ 
in a $proper^+$ form does not require CWA \cite{LiuLakemeyer2009}, one can 
develop a SC-based conformant planner that solves (\ref{planningReiter})
in this more general case, see \cite{BatusovMScThesis2014}. The latter planner 
does iterative deepening non-informed forward search to find a plan. Notice that 
if the initial theory $\dsz$ is in a $proper^+$ form, i.e., with the 
$\forall$-quantifier over object variables, then it is not clear how this planning 
instance can be formulated in PDDL \cite{DBLP:series/synthesis/2019Haslum}.
\cite{LiuLevesqueIJCAI2005} developed their tractable solution to the projection 
problem when the initial theory is proper, and mentioned they would like to 
develop a first-order planning system. This research direction led to
\cite{FanCaiLiLiuAAAI2012}, where 
$\dsz$ is given in 
a $proper^+$ form, and an implementation is tested on small instances of
Wumpus world and BW, but planning is assisted with Golog programs, and 
domain independent heuristics are not considered.

In fact, most of the previous research on deductive planning in SC centered 
around control strategies formulated in Golog, e.g., see 
\cite{BaierFritzMcIlraithICAPS2007} and earlier work reviewed in Chapter 12 of
\cite{GhallabNauTraverso2004}, where implementations rely on DCA and CWA. 
A related line of research, e.g., see  
\cite{ClassenEyerichLakemeyerNebelIJCAI2007,RogerHelmertNebelKR2008,ClassenRoegerLakemeyerNebelKI2012,ClassenPhD2013,RoegerPhDThesis2014}, 
explored semantics of planning languages in SC and possibilities of combining 
Golog-style planning in SC with an external heuristic PDDL planner.  But their research 
did not produce a deductive SC-based heuristic planner like our TPLH planner, their 
implementations implicitly required DCA, and it is not clear how their integration 
can be upgraded to infinite models with numerical fluents and continuous processes. 
We believe that our approach is more future-proof since it is more open to 
extensions, e.g., see \cite{BatusovSoutchanskiICAPS2019}. We have to mention also 
the previous work on planning with declarative, domain-specific heuristics, e.g., see 
\cite{BacchusKabanzaAI2000,ParmarASP2001WSh,AaratiParmarAAAI2002,Sierra-SantibanezAMAI2003,Sierra-SantibanezAI2004}, 
but our work is different since we developed a deductive lifted planner that 
controls  search with  a domain independent numerical heuristic.

More broadly, there are several well-known
best-first search (BFS) planners implemented in PROLOG, e.g., see
\cite{SterlingShapiro1994,PooleMackworthGoebel1998,Bratko2001}. However,
all these planners were doing search in a state space, not over situations, 
there were no connections with SC, and they did not have a domain
independent heuristic. 

As for answer set planning \cite{SonPontelliBalducciniSchaub2022}, this survey 
paper reviews the planners that have to perform grounding in advance, since 
otherwise they cannot take advantage of existing ASP solvers. Therefore, those 
planners are not lifted. Moreover, they are not doing BFS to compute a plan,
since there are no domain independent heuristics for answer set planning 
\cite{GebserEtAl-AAAI2013,DimopoulosGebserEtAl-TPLP2019}.

\cite{GhallabNauTraverso2004}, see Chapter 5, review early research on plan-space 
planning and several well-known systems, e.g.,
\cite{McAllesterRosenblittAAAI1991,PenberthyWeldKR1992,YounesSimmons2003,NguyenKambhampatiIJCAI2001}. 
None of these had any connections with SC or with deductive planning, but TPLH 
is designed from the first principles as theorem proving in SC. 
Note that only the plan-space search algorithms need a bound on the plan length
to guarantee completeness, but the modern model-based planners like FD and BFWS 
do not need it, since they search in the finite state space, but not over partial plans.
It remains to be seen if any of the methods from plan-space planning can help 
design a more efficient lifted deductive planner.

Most of the modern model-based 
planning systems are not lifted, i.e., they require
construction of a completely grounded transition system before search for a plan
can even start. Therefore, a non-lifted model-based approach has issues with 
scalability as the number of objects in the domain increases. There are several
realistic applications that demonstrate that grounding often results in 
out-of-memory problem, e.g., they are mentioned in \cite{CorreaAAAI22,MasoumiAntoniazziSoutchanski2015}. 
This limitation has been recently recognized by the planning community, and within 
model-based approach there are several competitive lifted planners, e.g., see 
\cite{LauerTorralbaFiserHollerWichlaczHoffmannIJCAI2021,CorreaFrancesPommereningHelmertICAPS2021,HorcikFiserTorralbaAAAI2022,CorreaSeipp2022,CorreaAAAI22}.
The main difference between our lifted approach and recent model-based lifted planners
is that we plan over situations that serve as concise pointers to large 
(potentially infinite) states, but other papers focus on model-based planning in
a finite state space. The latter approach has limits as the size of the state 
space can exceed available memory. If we used a lifted version of FF-heuristic,
as in  \cite{RidderFoxICAPS2014}, then our planner would still have minimal memory 
requirements even in the domains with continuous numerical fluents.

\section{Conclusion and Future Work}
We developed a sound and complete lifted planner based on theorem proving in the 
situation calculus. It does A$^*$ search for a plan in a tree of situations, but
not in a state space, and therefore it has minimal memory footprint. 
It is controlled using FF-inspired heuristic. To the best of our knowledge, 
TPLH is the first deductive planner based on SC with a domain independent heuristic.

In future, we would like to develop lifted versions of several heuristics, 
implement them efficiently, introduce tie-breaking for the cases when heuristic 
values are the same, consider lazy search and preferred actions. 
It did not escape from our attention that goal counting and novelty heuristics 
from BFWS work really well, and TPLH can benefit from these ideas. It is easy
to consider arbitrary action costs within TPLH. It is relatively easy to develop
a planner that works not only with context-free domains, but also with more 
general BATs, where SSAs have context conditions. 

  It happens that deductive planning in SC leads naturally to lifted planning 
with action schemas at run time. However, in this paper we do not compare our 
planner with other recent lifted planners. This study remains 
an interesting and important future work. 

Since we ground actions at run-time by evaluating their preconditions, and 
this is one of the computational bottlenecks, e.g., in the domains with complex 
preconditions \cite{MasoumiAntoniazziSoutchanski2015,QovaiziMScThesis2019}, 
we need a better algorithm for finding possible actions. 

In addition, we would like to develop an implementation that does not rely on 
DCA for objects, e.g., an implementation for the planning problems where the actions 
can create or destroy objects. This is doable within our theorem-proving approach 
to planning.

\section{Acknowledgments}
Thanks to the Natural Sciences and Engineering Research Council of Canada for
partial funding of this research.

\bibliographystyle{kr}

\bibliography{aaai23}

\end{document}